\documentclass[10pt,twocolumn,letterpaper]{article}

\usepackage{cvpr}
\usepackage{bm}
\usepackage{times}
\usepackage{epsfig}
\usepackage{graphicx}
\usepackage{amsmath}
\usepackage{amssymb}
\usepackage{multirow}
\usepackage{subfigure}
\usepackage{chngpage}
\usepackage{xcolor,colortbl}
\usepackage{bbold}
\usepackage{dsfont}
\usepackage{booktabs}
\usepackage{tabulary,overpic,xcolor}

\definecolor{citecolor}{RGB}{34,139,34}
\usepackage[pagebackref=true,breaklinks=true,letterpaper=true,colorlinks, citecolor=citecolor,bookmarks=false]{hyperref}

\usepackage[pagebackref=true,breaklinks=true,letterpaper=true,colorlinks,bookmarks=false]{hyperref}
\cvprfinalcopy % *** Uncomment this line for the final submission

\def\cvprPaperID{1247} % *** Enter the CVPR Paper ID here

\newcommand{\myparagraph}[1]{{\vspace{0.5em} \noindent \bf #1}}

\begin{document}

%%%%%%%%% TITLE
\title{Repulsion Loss: Detecting Pedestrians in a Crowd}

\author{Xinlong Wang$^1$\thanks{The work was done when Xinlong Wang and Tete Xiao were interns at Megvii, Inc.} \quad Tete Xiao$^{2*}$ \quad Yuning Jiang$^3$ \quad Shuai Shao$^3$ \quad Jian Sun$^3$ \quad Chunhua Shen$^4$ \\ \\
\and
$^1$Tongji University\\
{\tt\small 1452405wxl@tongji.edu.cn}\\
\and
$^2$Peking University\\
{\tt\small jasonhsiao97@pku.edu.cn}\\
\and
$^3$Megvii, Inc. \\
{\tt\small jyn, shaoshuai, sunjian@megvii.com}\\
\and
$^4$The University of Adelaide\\
{\tt\small chunhua.shen@adelaide.edu.au}\\
}
\maketitle

\begin{abstract}
Detecting individual pedestrians in a crowd remains a challenging problem since the pedestrians often gather together and occlude each other in real-world scenarios. In this paper, we first explore how a state-of-the-art pedestrian detector is harmed by crowd occlusion via experimentation, providing insights into the crowd occlusion problem. Then, we propose a novel bounding box regression loss specifically designed for crowd scenes, termed repulsion loss. This loss is driven by two motivations: the attraction by target, and the repulsion by other surrounding objects. The repulsion term prevents the proposal from shifting to surrounding objects thus leading to more crowd-robust localization. Our detector trained by repulsion loss outperforms the state-of-the-art methods with a significant improvement in occlusion cases.
\end{abstract}
%%%%%%%%%%%%%%%%%%%%%%%%%%%%%%%%%%%%%%%%%

%%%%%%%%%%%%%%%%%%%%%%%%%%%%%%%%%%%%%%%%%
\vspace{-0.2cm}
\section{Introduction}
\label{sec:intro}
%%%%%%%%%%%%%%%%%%%%%%%%%%%%%%%%%%%%%%%%%

Occlusion remains one of the most significant challenges in object detection although great progress has been made in recent years~\cite{Girshick_2014_CVPR,Girshick_2015_ICCV,NIPS2015_5638,Lin_2017_CVPR,cai2016unified,Lin_2017_ICCV,he2017mask,dai2016r}. In general, occlusion can be divided into two groups: {\it inter-class occlusion} and {\it intra-class occlusion}. The former one occurs when an object is occluded by stuff or objects of other categories, while the latter one, also referred to as {\it crowd occlusion}, occurs when an object is occluded by objects of the same category.

\begin{figure}[!htp]
\includegraphics[width=0.48\textwidth,height=0.24\textwidth]{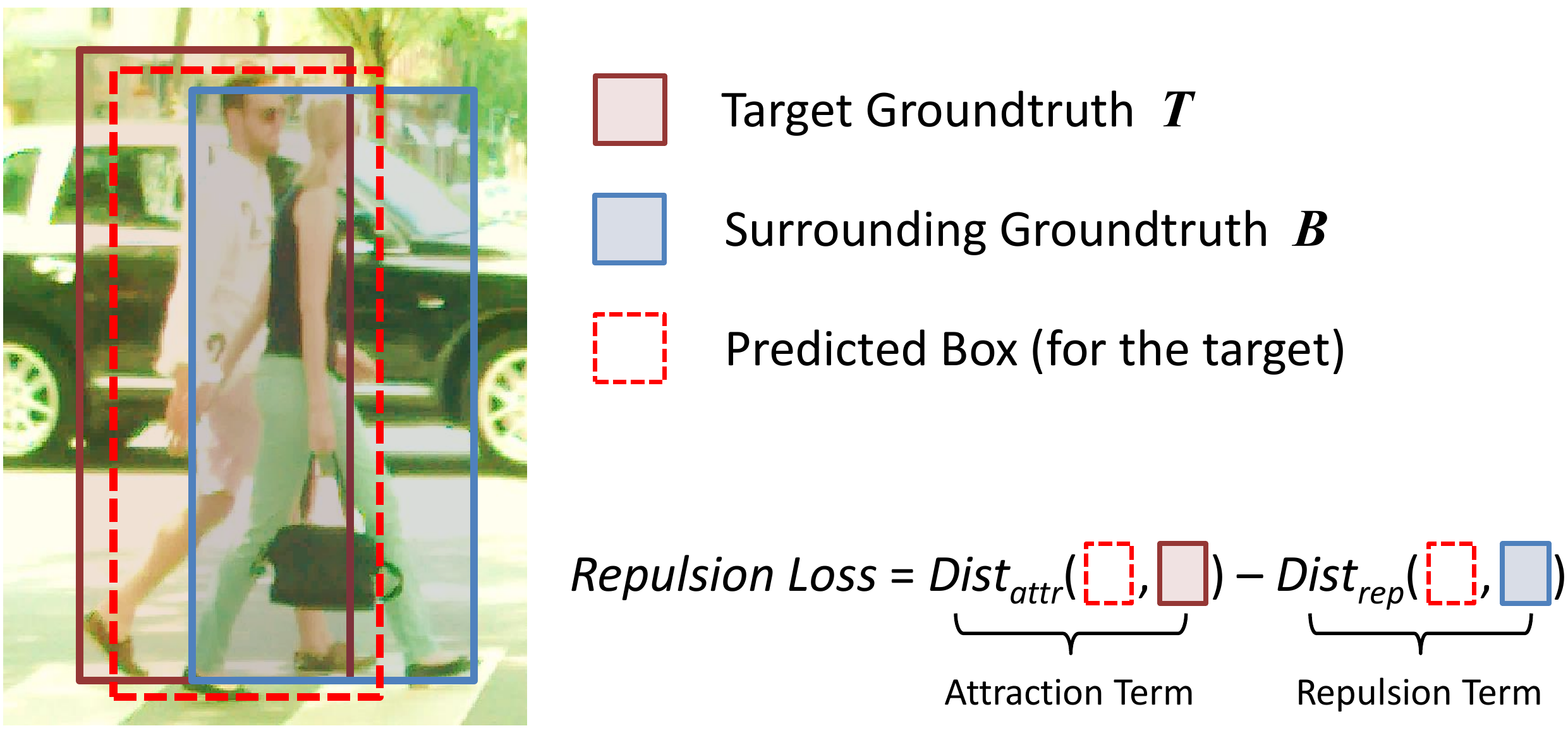}
\caption{Illustration of our proposed repulsion loss. The repulsion loss consists of two parts: the attraction term to narrow the gap between a proposal and its designated target, as well as the repulsion term to distance it from the surrounding non-target objects.}
\label{fig:intro}
\end{figure}

In pedestrian detection~\cite{zhang2016far,hosang2015taking,dollar2009integral,dollar2014fast,dollar2009pedestrian,mao2017can}, crowd occlusion constitutes the majority of occlusion cases. The reason is that in application scenarios of pedestrian detection, \eg, video surveillance and autonomous driving, pedestrians often gather together and occlude each other. For instance, in the CityPersons dataset~\cite{zhang2017citypersons}, there are a total of $3,157$ pedestrian annotations in the validation subset, among which $48.8\%$ of them overlap with another annotated pedestrian whose Intersection over Union (IoU) is above $0.1$. Moreover, $26.4\%$ of all pedestrians have considerable overlaps with another annotated pedestrian whose IoU is above $0.3$. The highly frequent crowd occlusion severely harms the performance of pedestrian detectors.

The main impact of crowd occlusion is that it significantly increases the difficulty in pedestrian localization. For example, when a target pedestrian $T$ is overlapped by another pedestrian $B$, the detector is apt to get confused since these two pedestrians have similar appearance features. As a result, the predicted boxes which should have bounded $T$ will probably shift to $B$, leading to inaccurate localization. Even worse, as the primary detection results are required to be further processed by non-maximum suppression (NMS), shifted bounding boxes originally from $T$ may be suppressed by the predicted boxes of $B$, in which $T$ turns into a missed detection. That is, crowd occlusion makes the detector sensitive to the threshold of NMS: a higher threshold brings in more false positives while a lower threshold leads to more missed detections. Such undesirable behaviors can harm most instance segmentation frameworks~\cite{he2017mask,li2017fully}, since they also require accurate detection results. Therefore, how to robustly localize each individual person in crowd scenes is one of the most critical issues for pedestrian detectors.

In state-of-the-art detection frameworks~\cite{Girshick_2015_ICCV,NIPS2015_5638,dai2016r,Lin_2017_CVPR}, the bounding box regression technique is employed for object localization, in which a regressor is trained to narrow the gap between proposals and ground-truth boxes measured by some kind of distance metrics (\eg, $\mathrm{Smooth}_{L1}$ or IoU). Nevertheless, existing methods only require the proposal to get close to its designated target, without taking the surrounding objects into consideration. As shown in Figure~\ref{fig:intro}, in the standard bounding box regression loss, there is no additional penalty for the predicted box when it shifts to the surrounding objects. This observation makes one wonder {\it whether the locations of its surrounding objects could be taken into account if we want to detect a target in a crowd?}

Inspired by the characteristics of a magnet, \ie, {\it magnets attract and repel}, in this paper we propose a novel localization technique, referred to as repulsion loss (RepLoss). With RepLoss, each proposal is required not only to approach its designated target $T$, but also to keep away from the other ground-truth objects as well as the other proposals whose designated targets are not $T$. In other words, the bounding box regressor with RepLoss is driven by two motivations: attraction by the target and repulsion by other surrounding objects and proposals. For example, as demonstrated in Figure~\ref{fig:intro}, the red bounding box shifting to $B$ will be given an additional penalty since it overlaps with a surrounding non-target object. Thus, RepLoss can prevent the predicted bounding box from shifting to adjacent overlapped objects effectively, which makes the detector more robust to crowd scenes. Our main contributions are as follows:

\begin{itemize}
\setlength{\itemsep}{0pt}
\setlength{\parskip}{0pt}
\setlength{\parsep}{0pt}
  \item We first experimentally study the impact of crowd occlusion on pedestrian detection. Specifically, on the CityPersons benchmark~\cite{zhang2017citypersons} we analyze both false positives and missed detections caused by crowd occlusion quantitatively, which provides important insights into the crowd occlusion problem.
  \item Two types of repulsion losses are proposed to address the crowd occlusion problem, namely RepGT Loss and RepBox Loss. RepGT Loss directly penalizes the predicted box for shifting to the other ground-truth objects, while RepBox Loss requires each predicted box to keep away from the other predicted boxes with different designated targets, making the detection results less sensitive to NMS.
  \item With the proposed repulsion losses, a crowd-robust pedestrian detector is trained end-to-end, which outperforms all the state-of-the-art methods on both CityPerson and Caltech-USA benchmarks~\cite{dollar2009pedestrian}. It should also be noted that the detector with repulsion loss significantly improves the detection accuracy for occlusion cases, highlighting the effectiveness of repulsion loss. Furthermore, our experiments on the PASCAL VOC~\cite{everingham2010pascal} detection dataset show that the RepLoss is also beneficial for general object detection, besides pedestrians. 
\end{itemize}

%%%%%%%%%%%%%%%%%%%%%%%%%%%%%%%%%%%%%%%%%
\section{Related Work}
%%%%%%%%%%%%%%%%%%%%%%%%%%%%%%%%%%%%%%%%%

\myparagraph{Object Localization.} With the recent development of convolutional neural networks (CNNs)~\cite{krizhevsky2012imagenet,simonyan2014very,he2016deep}, great progress has been made in object detection, in which object localization is generally framed as a regression problem that relocates an initial proposal to its designated target. In R-CNN~\cite{Girshick_2014_CVPR}, a linear regression model is trained with respect to the Euclidean distance of coordinates of a proposal and its target. In~\cite{Girshick_2015_ICCV}, the $\mathrm{Smooth}_{L1}$ Loss is proposed to replace the Euclidean distance used in R-CNN for bounding box regression. \cite{NIPS2015_5638} proposes the region proposal network (RPN), in which bounding box regression is performed twice to transform predefined anchors into final detection boxes. Densebox~\cite{huang2015densebox} proposes an anchor-free, fully convolutional detection framework. IoU Loss is proposed in~\cite{yu2016unitbox} to maximize the IoU between a ground-truth box and a predicted box.  We note that a method proposed by Desai \etal~\cite{desai2011discriminative} also exploits  the attraction and repulsion between objects to capture the spatial arrangements of various object classes, still, it is to address the problem of object classification via a global model. In this work, we will demonstrate the effectiveness of the Repulsion Loss for object localization in crowd scenes.

\myparagraph{Pedestrian Detection.} Pedestrian detection is the first and an critical step for many real-world applications. Traditional pedestrian detectors, such as \verb'ACF'~\cite{dollar2014fast}, \verb'LDCF'~\cite{nam2014local} and \verb'Checkerboard'~\cite{zhang2015filtered}, exploit various filters on Integral Channel Features (\verb'IDF')~\cite{dollar2009integral} with sliding window strategy to localize each target. Recently, the CNN-based detectors~\cite{li2017scale,zhang2016faster,mao2017can,hosang2015taking,yang2015convolutional} show great potential in dominating the field of pedestrian detection. In~\cite{yang2015convolutional,zhang2016faster}, features from a Deep Neural Network rather than hand-crafted features are fed into a boosted decision forest. \cite{mao2017can} proposes a multi-task trained network to further improve detection performance. Also in~\cite{ouyang2012discriminative,tian2015deep,zhou2017multi}, a part-based model is utilized to handle occluded pedestrians. \cite{Hosang_2017_CVPR} works on improving the robustness of NMS, but it ends up relying on an additional network for post-processing. In fact, few of previous works focus on studying and overcoming the impact of crowd occlusion.

%%%%%%%%%%%%%%%%%%%%%%%%%%%%%%%%%%%%%%%%%
\section{What is the Impact of Crowd Occlusion?}
\label{sec:howmuchanalysis}
%%%%%%%%%%%%%%%%%%%%%%%%%%%%%%%%%%%%%%%%%

To provide insights into the crowd occlusion problem, in this section, we experimentally study how much crowd occlusion influences pedestrian detection results. Before delving into our analysis, first we introduce the dataset and the baseline detector that we use.

\subsection{Preliminaries}
\myparagraph{Dataset and Evaluation Metrics.}
CityPersons~\cite{zhang2017citypersons} is a new pedestrian detection dataset on top of the semantic segmentation dataset CityScapes~\cite{cordts2016cityscapes}, of which $5,000$ images are captured in several cities in Germany. A total of ${\sim}35,000$ persons with an additional ${\sim}13,000$ ignored regions, both bounding box annotation of all persons and annotation of visible parts are provided. All of our experiments involved CityPersons are conducted on the {\it reasonable} train/validation sets for training and testing, respectively. For evaluation, the log miss rate is averaged over the false positive per image (FPPI) range of $[10^{-2},~10^{0}]$ ($\mathrm{MR}^{-2}$) is used (lower is better).

\myparagraph{Detector.}
Our baseline detector is the commonly used Faster R-CNN~\cite{NIPS2015_5638} detector modified for pedestrian detection, generally following the settings in Zhang \etal~\cite{zhang2016far} and Mao \etal~\cite{mao2017can}. The difference between our implementation and theirs is that we replace the VGG-16 backbone with the faster and lighter ResNet-50~\cite{he2016deep} network. It is worth noting that ResNet is rarely used in pedestrian detection, since the down-sampling rate at convolution layers is too large for the network to detect and localize small pedestrians. To handle this, we use dilated convolution and the final feature map is $1/8$ of input size. The ResNet-based detector achieves $14.6$ $\mathrm{MR}^{-2}$ on the validation set, which is sightly better than the reported result ($15.4$ $\mathrm{MR}^{-2}$) in~\cite{zhang2017citypersons}.

\begin{figure}[t]
\includegraphics[width=0.485\textwidth,height=0.3\textwidth]{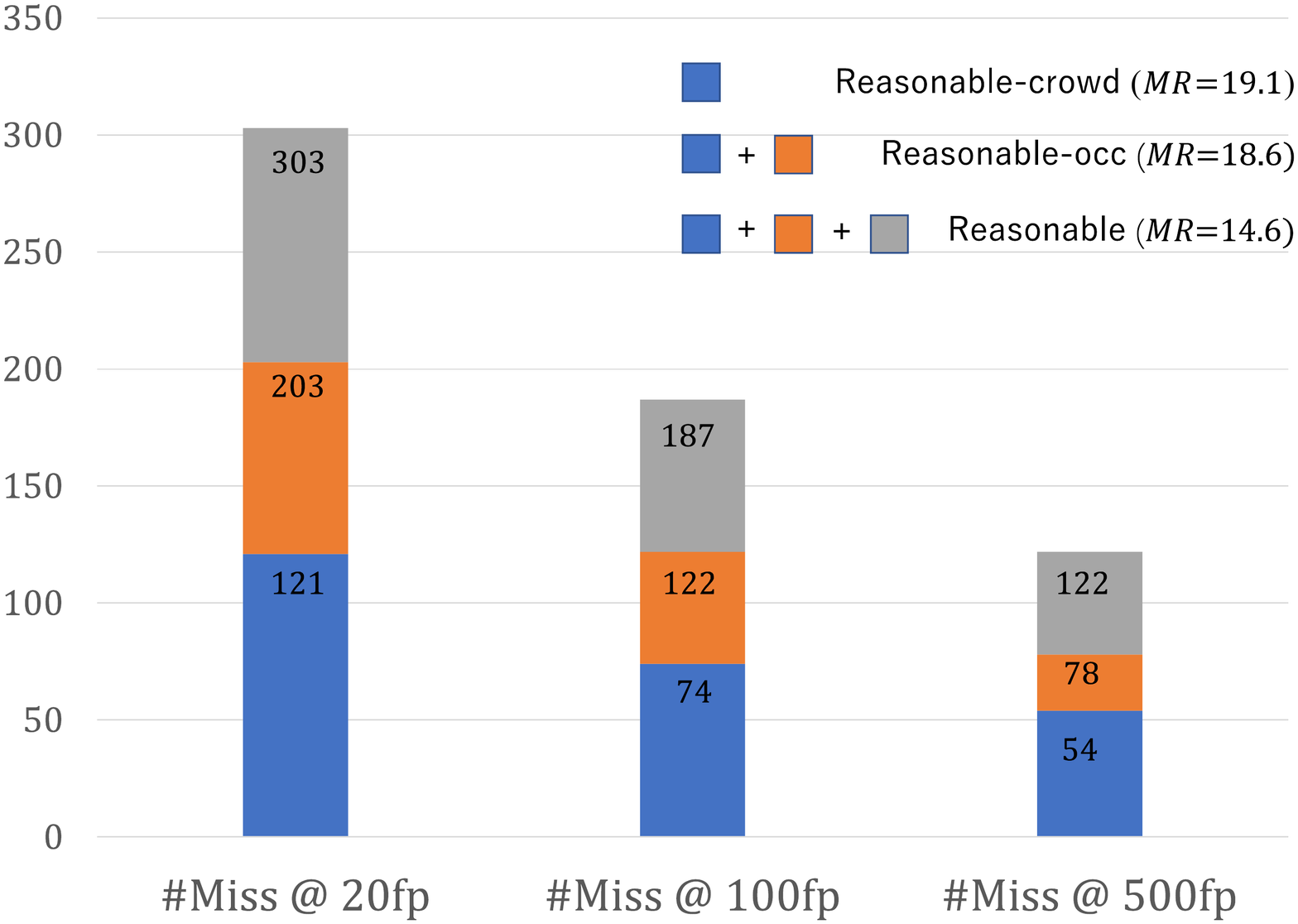}
\caption{Missed detection numbers and $\mathrm{MR}^{-2}$ scores of our baseline on the reasonable, reasonable-occ, reasonable-crowd subsets. Of all missed detection in reasonable-occ subset, crowd occlusion accounts for ${\sim}60\%$, making it a main obstacle for addressing occlusion issues.}
\label{fig:missdetection}
\end{figure}

\subsection{Analysis on Failure Cases}

\begin{figure}[!tbp]
\centering
\subfigure[]{
\includegraphics[width=0.38\textwidth]{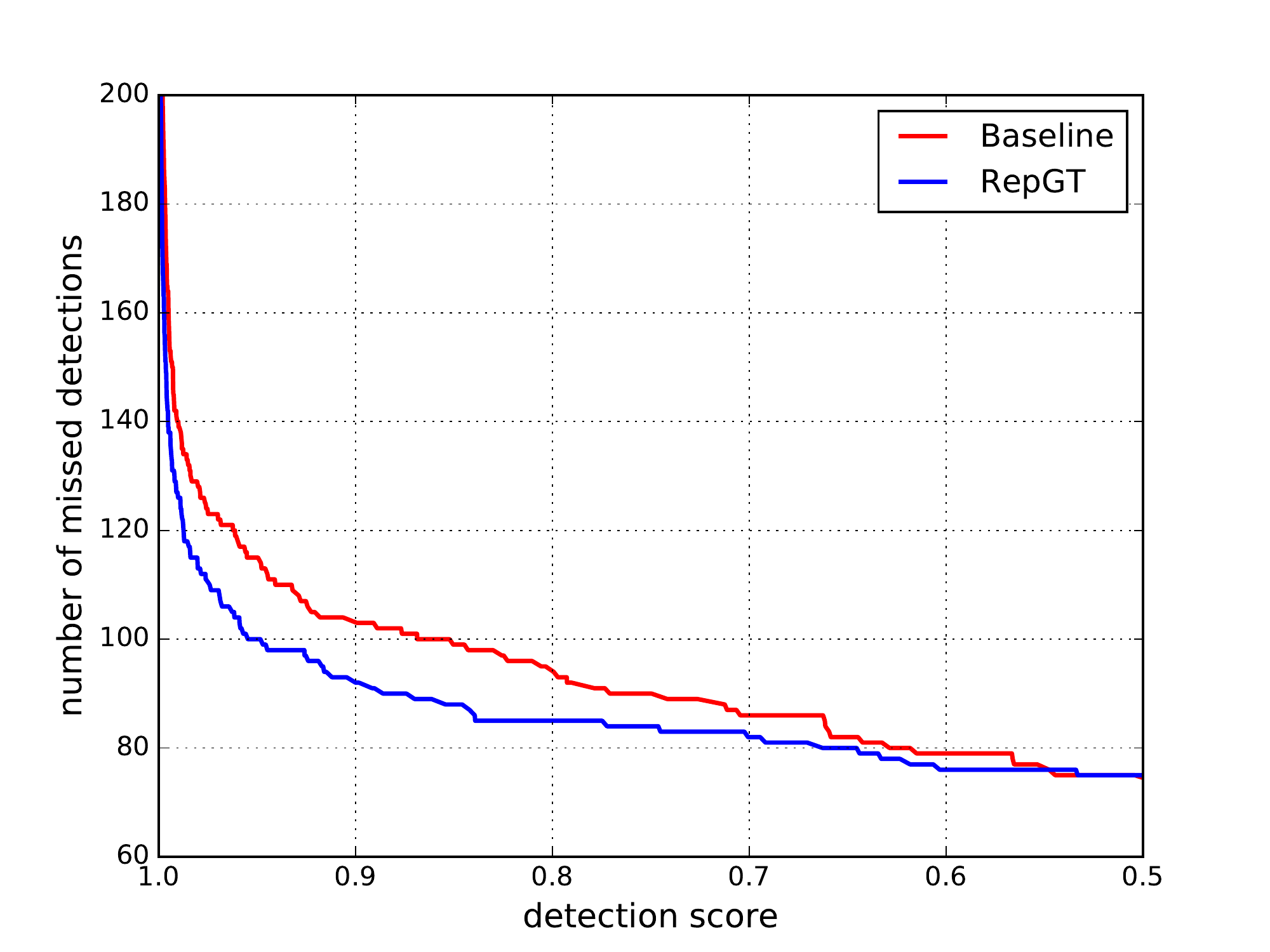}
\label{fig:crowd_miss}
}
\subfigure[]{
\includegraphics[width=0.38\textwidth]{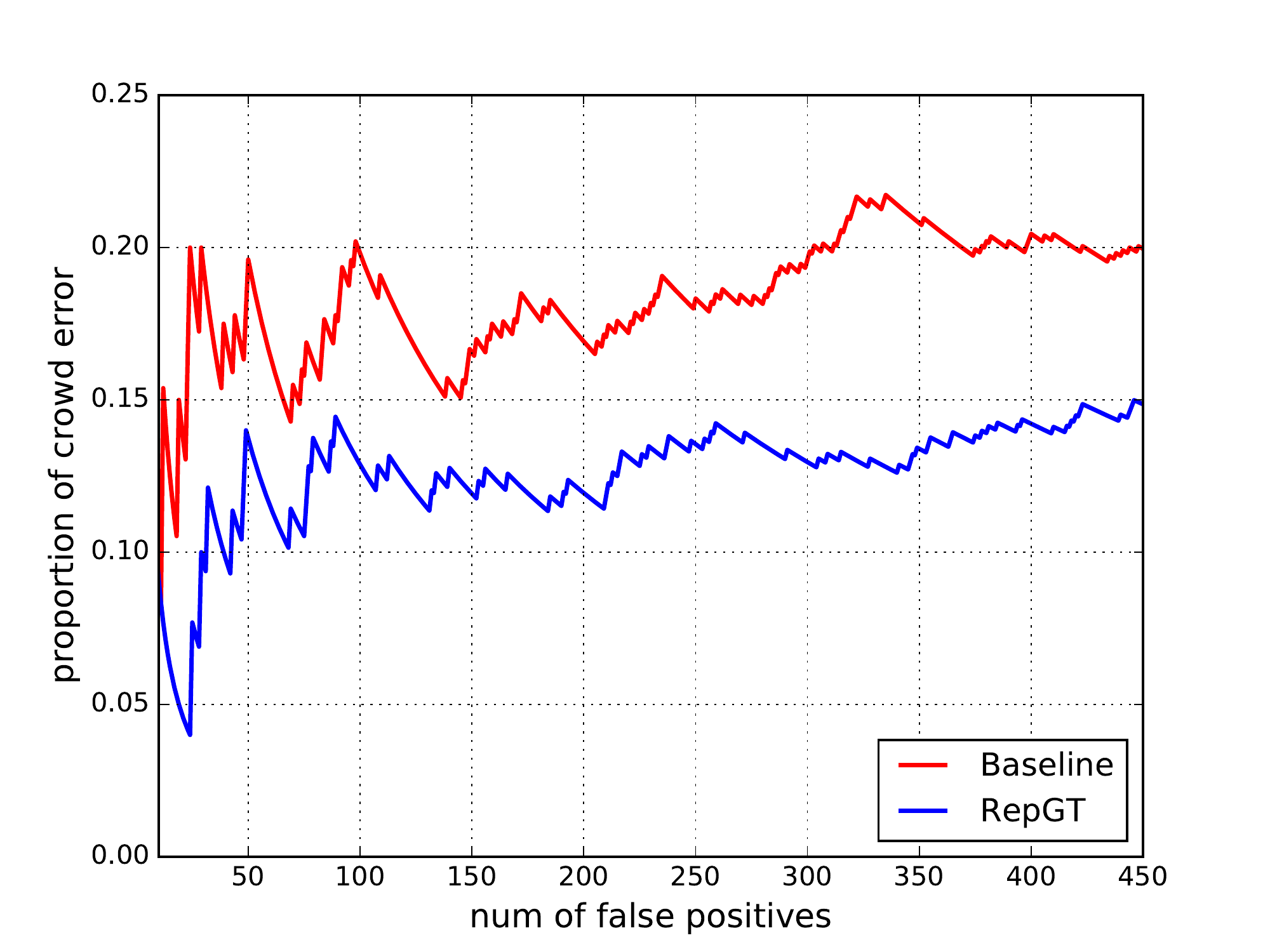}
\label{fig:crowd_fp}
}
\caption{Errors analysis of our baseline and RepGT. (a) The number of missed detections in reasonable-crowd subset under different detection scores. (b) The proportion of false positives caused by crowd occlusion of all false positives. RepGT Loss effectively reduces missed detections and false positives caused by crowd occlusion.}
\vspace{-0.2cm}
\label{fig:crowderrors}
\end{figure}

\myparagraph{Missed Detections.}
With the results of the baseline detector, we first analyze missed detections caused by crowd occlusion. Since the bounding box annotation of the visible part of each pedestrian is provided in CityPersons, the occlusion can be calculated as $occ \triangleq 1 - \frac{area(BBox_{visible})}{area(BBox)}$. We define a ground-truth pedestrian whose $occ \geq 0.1$ as an occlusion case, and one whose $occ \geq 0.1$ and $IoU \geq 0.1$ with any other annotated pedestrian as a crowd occlusion case. By definition, from the total $1,579$ non-ignored pedestrian annotations in the reasonable validation set, two subsets are extracted: the {\it reasonable-occ} subset, consisting of $810$ occlusion cases ($51.3\%$) and the {\it reasonable-crowd} subset, consisting of $479$ crowd occlusion cases ($30.3\%$). Obviously the reasonable-crowd subset is also a subset of reasonable-occ subset.

In Figure~\ref{fig:missdetection}, we report the numbers of missed detections and $\mathrm{MR}^{-2}$ on the reasonable, reasonable-occ and reasonable-crowd subsets. We observe that the performance drops significantly from $14.6$ $\mathrm{MR}^{-2}$ on the reasonable set to $18.6$ $\mathrm{MR}^{-2}$ on the reasonable-occ subset; of all missed detections at 20, 100, and 500 false positives, occlusion makes up approximately $60\%$, indicating that it is a main factor which harms the performance of the baseline detector. Of missed detections in the reasonable-occ subset, the proportion of crowd occlusion stands at nearly $60\%$, making it a main obstacle for addressing occlusion issues in pedestrian detection. Moreover, the miss rate on the reasonable-crowd subset ($19.1$) is even higher than the reasonable-occ subset ($18.6$), indicating that crowd occlusion is an even harder problem than inter-class occlusions; when we lower the threshold from 100 to 500 false positives, the portion of missed detections caused by crowd occlusion becomes larger (from $60.7\%$ to $69.2\%$). It implies that missed detections caused by crowd occlusion are hard to be rescued by lowering the threshold.

In Figure~\ref{fig:crowd_miss}, the red line shows how many ground-truth pedestrians are missed in the reasonable-crowd subset with different detection scores. As in real-world applications, only predicted bounding boxes with high confidence will be considered, the large number of missed detections on the top of the curve implies we are far from saturation for real-world applications.

\myparagraph{False Positives.}
We also analyze how many false positives are caused by crowd occlusion. We cluster all false positives into three categories: background, localization and crowd error. A background error occurs when a predicted bounding box has $IoU < 0.1$ with any ground-truth pedestrian, while a localization error has $IoU \geq 0.1$ with only one ground-truth pedestrian. Crowd errors are those who have $IoU \geq 0.1$ with at least two ground-truth pedestrians.

\begin{figure}[t]
\includegraphics[width=0.485\textwidth]{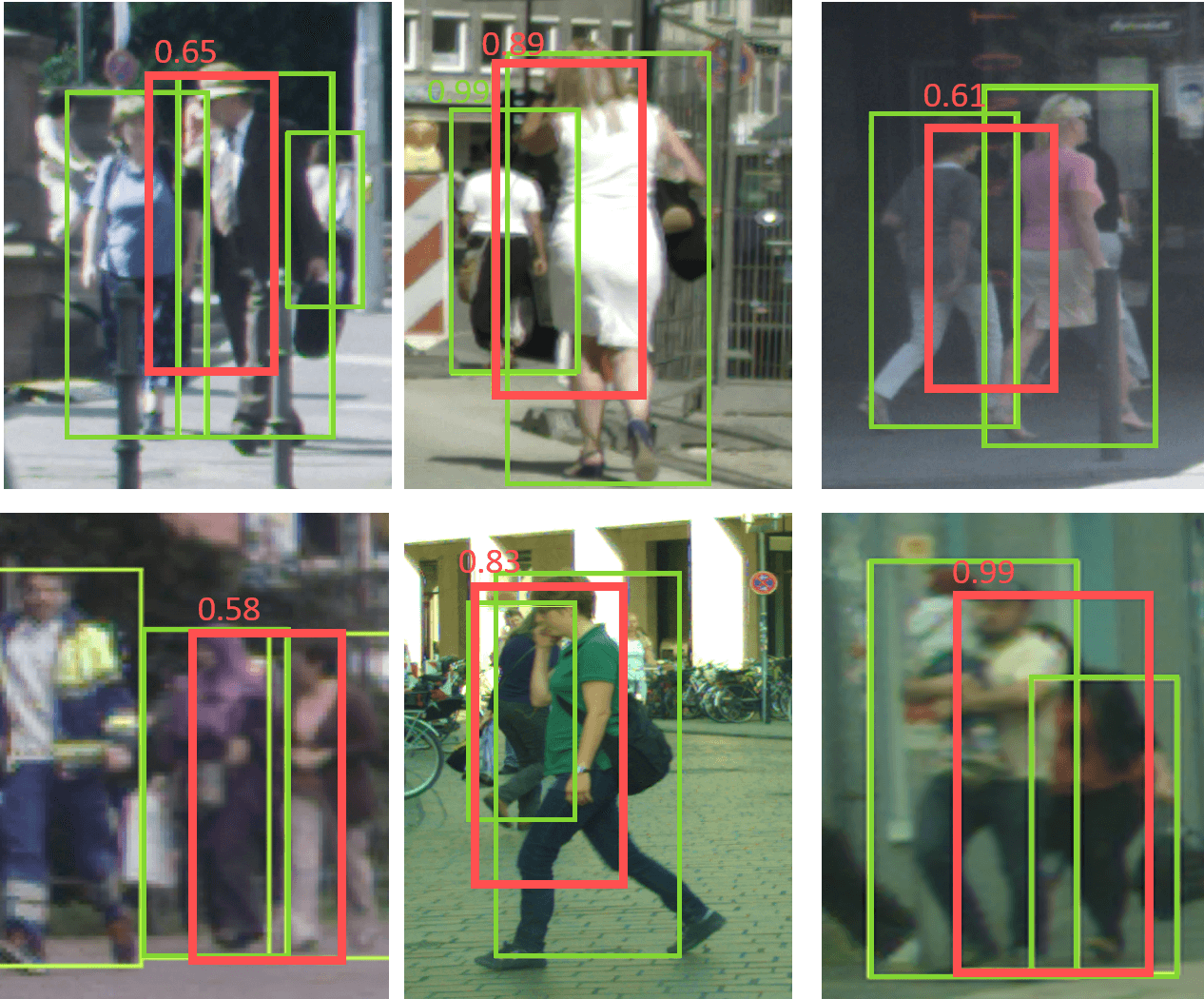}
\caption{The visualization examples of the crowd errors. Green boxes are correct predicted bounding boxes, while red boxes are false positives caused by crowd occlusion. The confidence scores outputted by detectors are also attached. The errors usually occur when a predict box shifts slightly or dramatically to neighboring ground-truth object (\eg, top-right one), or bounds the union of several overlapping ground-truth objects (\eg, bottom-right one).}
\vspace{-0.2cm}
\label{fig:fp_vis}
\end{figure}

After that we count the number of crowd errors and calculate its proportion of all false positives. The red line in Figure~\ref{fig:crowd_fp} shows that crowd errors contribute to a relative large proportion (about $20\%$) of all false positives. Through visualization in Figure~\ref{fig:fp_vis}, we observe that the crowd errors usually occur when a predict box shifts slightly or dramatically to neighboring non-target ground-truth objects, or bounds the union of several overlapping ground-truth objects together. Moreover, the crowd errors usually have relatively high confidences thus leading to top-ranked false positives. It indicates that to improve the robustness of detectors to crowd scenes, more discriminative loss is needed when performing bounding box regression. More visualization examples can be found in supplementary material.

\myparagraph{Conclusion.}
The analysis on failure cases validates our observation: pedestrian detectors are surprisingly tainted by crowd occlusion, as it constitutes the majority of missed detections and results in more false positives by increasing the difficulty in localization. To address these issues, in Section~\ref{sec:approach}, the repulsion loss is proposed to improve the robustness of pedestrian detectors to crowd scenes.

%%%%%%%%%%%%%%%%%%%%%%%%%%%%%%%%%%%%%%%%%
\section{Repulsion Loss}
\label{sec:approach}
%%%%%%%%%%%%%%%%%%%%%%%%%%%%%%%%%%%%%%%%%

In this section we introduce the repulsion loss to address the crowd occlusion problem in detection. Inspired by the characteristics of magnet, \ie, {\it magnets attract and repel}, the Repulsion Loss is made up of three components, defined as:
\begin{equation}
L=L_\mathrm{Attr} + \alpha * L_\mathrm{RepGT} + \beta * L_\mathrm{RepBox},
\end{equation}
where $L_\mathrm{Attr}$ is the {\it attraction} term which requires a predicted box to approach its designated target, while $L_\mathrm{RepGT}$ and $L_\mathrm{RepBox}$ are the {\it repulsion} terms which require a predicted box to keep away from other surrounding ground-truth objects and other predicted boxes with different designated targets, respectively. Coefficients $\alpha$ and $\beta$ act as the weights to balance auxiliary losses.

For simplicity we consider only two-class detection in the following, assuming all ground-truth objects are from the same category. Let $P =(l_P, t_P, w_P, h_P)$ and $G =(l_G, t_G, w_G, h_G)$ be the proposal bounding box and ground-truth bounding box which are represented by their coordinates of left-top points as well as their widths and heights, respectively. $\mathcal{P_+} = \{P\}$ is the set of all positive proposals (those who have a high IoU (\eg, $IoU \geq 0.5$) with at least one ground-truth box are regarded as positive samples, while negative samples otherwise), and $\mathcal{G} = \{G\}$ is the set of all ground-truth boxes in one image.

\myparagraph{Attraction Term.} With the objective to narrow the gap between predicted boxes and ground-truth boxes measured by some kind of distance metrics\footnote{Here the distance is simply a measurement of difference of two bounding boxes. It may not satisfy triangle inequality.}, \eg, Euclidean distance~\cite{Girshick_2014_CVPR}, $\mathrm{Smooth}_{L1}$ distance~\cite{Girshick_2015_ICCV} or IoU~\cite{yu2016unitbox}, attraction loss has been commonly adopted in existing bounding box regression techniques. To make a fair comparison, in this paper we adopt $\mathrm{Smooth}_{L1}$ distance for the attraction term as in~\cite{mao2017can,zhang2017citypersons}. We set smooth parameter in $\mathrm{Smooth}_{L1}$ as $2$. Given a proposal $P \in \mathcal{P_+}$, we assign the ground-truth box who has the maximum IoU as its designated target: $G_{Attr}^P = \arg\max_{G \in \mathcal{G}} IoU(G, P)$. $B^P$ is the predicted box regressed from proposal $P$. Then the attraction loss could be calculated as:

\begin{equation}
L_\mathrm{Attr} = \frac{\sum_{P \in \mathcal{P_+}} \mathrm{Smooth}_{L1}(B^P, G_{Attr}^{P})}{|\mathcal{P_+}|}.
\label{eqn:attr}
\end{equation}

\begin{figure}[t]
\centering
\includegraphics[width=0.38\textwidth]{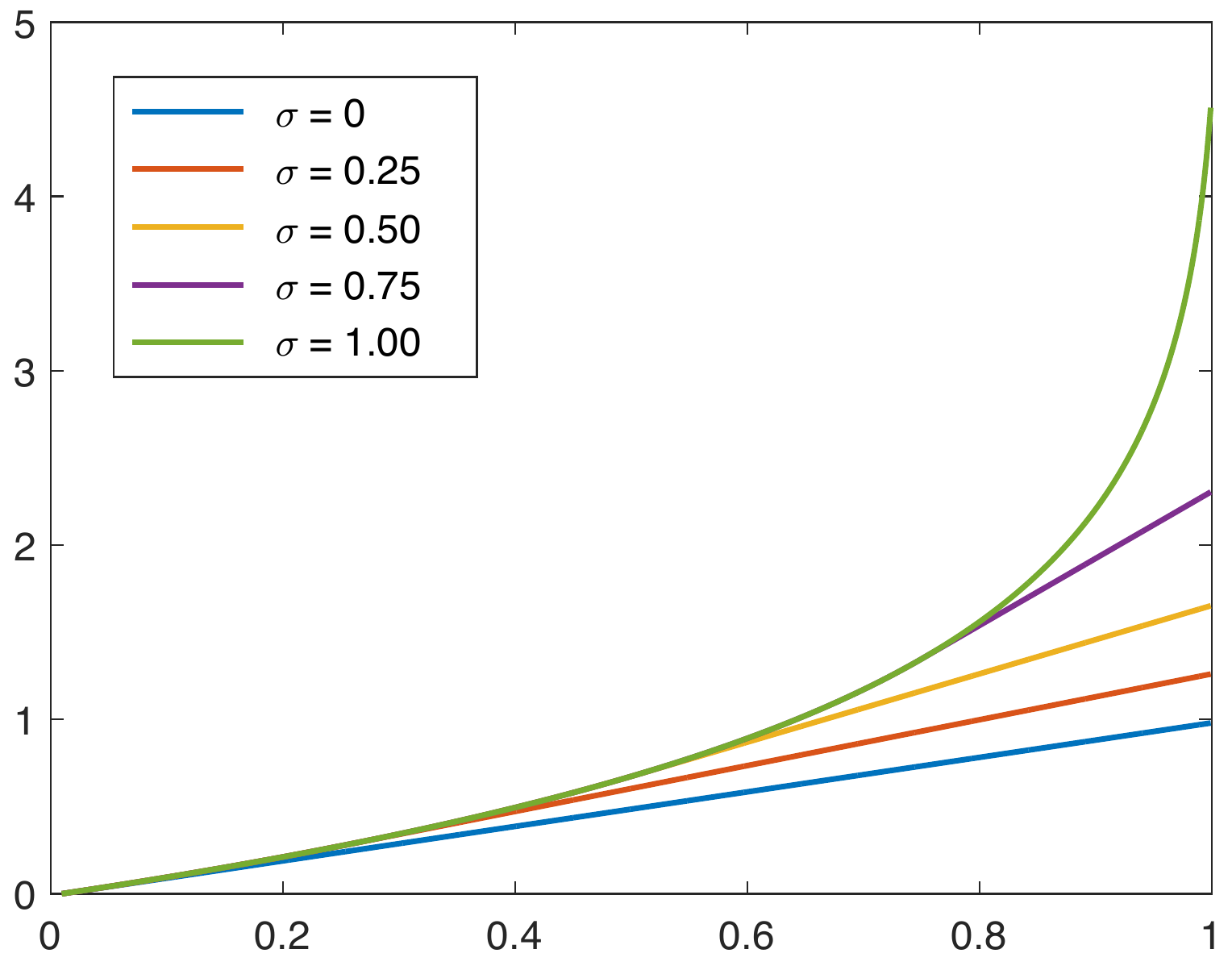}
\caption{The curves of $\mathrm{Smooth}_{ln}$ under different smooth parameter $\sigma$. The smaller $\sigma$ is, the less sensitive loss is to the outliers.}
\vspace{-0.2cm}
\label{fig:smoothparam}
\end{figure}

\myparagraph{Repulsion Term (RepGT).} The RepGT Loss is designed to repel a proposal from its neighboring ground-truth objects which are not its target. Given a proposal $P \in \mathcal{P_+}$, its repulsion ground-truth object is defined as the ground-truth object with which it has the largest IoU region except its designated target:
\begin{equation}
G_{Rep}^P = \mathop{\arg\max}_{G \in \mathcal{G}\setminus\{G_{Attr}^P\}} IoU(G, P).
\end{equation}
Inspired by IoU Loss in~\cite{yu2016unitbox}, the RepGT Loss is calculated to penalize the overlap between $B^P$ and $G_{Rep}^P$.  The overlap between $B^P$ and $G_{Rep}^P$ is defined by Intersection over {\it Ground-truth} (IoG): $IoG(B, G) \triangleq \frac{area(B \cap G)}{area(G)}$. As $IoG(B, G) \in [0,1]$, we define RepGT Loss as:
\begin{equation}
L_\mathrm{RepGT} = \frac{\sum_{P \in \mathcal{P_+}} \mathrm{Smooth}_{ln}\left(IoG(B^P, G_{Rep}^P)\right)}{|\mathcal{P_+}|},
\label{eqn:repgt}
\end{equation}
where
\begin{equation}
\mathrm{Smooth}_{ln} = \left \{
\begin{array}{lr}
    \vspace{0.2cm}
    -\ln{(1-x)}  \quad & x \leq \sigma \\
    \dfrac{x-\sigma}{1-\sigma} - \ln{(1-\sigma)} \quad & x > \sigma \\
\end{array}
\right.
\label{eqn:smoothln}
\end{equation}
is a smoothed $\ln$ function which is continuously differentiable in $(0,1)$, and $\sigma \in [0,1)$ is the smooth parameter to adjust the sensitiveness of the repulsion loss to the outliers. Figure~\ref{fig:smoothparam} shows its curve with different $\sigma$. From Eqn.~\ref{eqn:repgt} and Eqn.~\ref{eqn:smoothln} we can see that the more a proposal tends to overlap with a non-target ground-truth object, a larger penalty will be added to the bounding box regressor by the RepGT Loss. In this way, the RepGT Loss could effectively stop a predicted bounding box from shifting to its neighboring objects which are not its target.

\myparagraph{Repulsion Term (RepBox).} NMS is a necessary post-processing step in most detection frameworks to merge the primary predicted bounding boxes which are supposed to bound the same object. However, the detection results will be affected significantly by NMS especially for the crowd cases. To make the detector less sensitive to NMS, we further propose the RepBox Loss whose objective is to repel each proposal from others with different designated targets. We divide the proposal set $\mathcal{P_+}$ into $|\mathcal{G}|$ mutually disjoint subsets based on the target of each proposal: $\mathcal{P_+} = \mathcal{P}_1 \cap \mathcal{P}_2 \cap \ldots \cap \mathcal{P}_{|\mathcal{G}|}$. Then for two proposals randomly sampled from two different subsets, $P_i \in \mathcal{P}_i$ and $P_j \in \mathcal{P}_j$ where $i,j=1,2,\ldots,{|\mathcal{G}|}$ and $i \neq j$, we expect that the overlap of predicted box $B^{P_i}$ and $B^{P_j}$ will be as small as possible. Therefore, the RepBox Loss is calculated as:
\begin{equation}
L_\mathrm{RepBox} = \frac{\sum_{i \neq j}{\mathrm{Smooth}_{ln}\left(IoU(B^{P_i}, B^{P_j})\right)}}{\sum_{i \neq j}{\mathds{1}{\left[IoU(B^{P_i}, B^{P_j}) > 0\right]}} + \epsilon},
\label{eqn:repbox}
\end{equation}
where $\mathds{1}$ is the identity function and $\epsilon$ is a small constant in case divided by zero. From Eqn.~\ref{eqn:repbox} we can see that to minimize the RepBox Loss, the IoU region between two predicted boxes with different designated targets needs to be small. That means, the RepBox Loss is able to reduce the probability that the predicted bounding boxes with different regression targets are merged into one after NMS, which makes the detector more robust to the crowd scenes.

\subsection{Discussion}
\myparagraph{Distance Metric.} It is worth noting that we choose the IoG or IoU rather than $\mathrm{Smooth}_{L1}$ metric to measure the distance between two bounding boxes in the repulsion term. The reason is that the values of IoG and IoU are bounded in range $[0,1]$ while $\mathrm{Smooth}_{L1}$ metric is boundless, \ie, if we use $\mathrm{Smooth}_{L1}$ metric in the repulsion term, in the RepGT Loss for example, it will require the predicted box to keep away from its repulsion ground-truth object as far as possible. On the contrary, IoG criteria only requires the predicted box to minimize the overlap with its repulsion ground-truth object, which better fits our motivation.

In addition, IoG is adopted in RepGT Loss rather than IoU because, in the IoU-based loss, the bounding box regressor may learn to minimize the loss by simply enlarging the bounding box size to increase the denominator $area(B^P \cup G_{Rep}^P)$. Therefore, we choose IoG whose denominator is a constant for a particular ground-truth object to minimize the overlap $area(B^P \cap G_{Rep}^P)$ directly.

\myparagraph{Smooth Parameter $\sigma$.}
Compared to~\cite{yu2016unitbox} which directly uses $-\ln(IoU)$ as loss function, we introduce a smoothed $\ln$ function $\mathrm{Smooth}_{ln}$ and a smooth parameter $\sigma$ in both RepGT Loss and RepBox Loss. As shown in Figure~\ref{fig:smoothparam}, we can adjust the sensitiveness of the repulsion loss to the outliers (the pair of boxes with large overlap) by the smooth parameter $\sigma$. Since the predicted boxes are much denser than the ground-truth boxes, a pair of two predicted boxes are more likely to have a larger overlap than a pair of one predicted box and one ground-truth box. It means that there will be more outliers in RepBox than in RepGT. So, intuitively, RepBox Loss should be less-sensitive to outliers (with small $\sigma$) than RepGT Loss. More detailed studies about the smooth parameter $\sigma$ as well as the auxiliary loss weights $\alpha$ and $\beta$ are provided Section~\ref{subsec:ablationexperiments}.

\section{Experiments}
\label{sec:exp}
The experiment section is organized as follows: we first introduce the basic experiment settings as well as the implementation details of repulsion loss in Section~\ref{subsec:implementation_details}; then the proposed RepGT Loss and RepBox Loss are evaluated and analyzed on the CityPersons~\cite{zhang2017citypersons} benchmark respectively in Section~\ref{subsec:ablationexperiments}; finally, in Section~\ref{subsec:compareison_with_stateoftheart}, the detector with repulsion loss is compared with the state-of-the-art methods side-by-side on both CityPersons~\cite{zhang2017citypersons} and Caltech-USA~\cite{dollar2009pedestrian}.

\subsection{Experiment Settings}
\label{subsec:implementation_details}
\myparagraph{Datasets.} Besides the CityPersons~\cite{zhang2017citypersons} benchmark introduced in Section~\ref{sec:howmuchanalysis}, we also carry out experiments on the Caltech-USA dataset~\cite{dollar2009pedestrian}. As one of several predominant datasets and benchmarks for pedestrian detection, Caltech-USA has witnessed inspiring progress in this field.
A total of 2.5-hour video is divided into training and testing subsets with $42,500$ frames and $4,024$ frames respectively. In ~\cite{zhang2016far}, Zhang \etal provide refined annotations, in which training data are refined automatically while testing data are meticulously re-annotated by human annotators. We conduct all experiments related to Caltech-USA on the new annotations unless otherwise stated.

\myparagraph{Training Details.} Our framework is implemented on our self-built fast and flexible deep learning platform. We train the network for $80k$ iterations and $160k$ iterations, with the base learning rate set to 0.016 and decreased by a factor of 10 after the first $60k$ and $120k$ iterations for CityPersons and Caltech-USA, respectively. The Stochastic Gradient Descent (SGD) solver is adopted to optimize the network on 4 GPUs. A mini-batch involves 1 image per GPU. Weight decay and momentum are set to $0.0001$ and $0.9$. Multi-scale training/testing are not applied to ensure fair comparisons with previous methods. For Caltech-USA, we use the $10$x set (${\sim}42k$ frames) for training. Online Hard Example Mining (OHEM)~\cite{shrivastava2016training} is used to accelerate convergence.

% combination of RepGT and RepBox

\begin{table}[!tbp]
\begin{center}
\setlength{\tabcolsep}{5pt}
\begin{tabular}{|c|c|c|c|c|c|c|}
\hline
  & \multicolumn{3}{c|}{$\mathrm{MR}^{-2}$} & \multicolumn{3}{c|}{Improvement} \\
\hline
$\sigma$ & 0 & 0.5 & 1.0 & 0 & 0.5 & 1.0 \\
\hline
RepGT & 14.3 & 14.5 & {\bf 13.7} & +0.3 & +0.1 & {\bf +0.9}  \\
\hline
RepBox & {\bf 13.7} & 14.2 & 14.3 & {\bf +0.9} & +0.4 & +0.3  \\
\hline
\end{tabular}
\end{center}
\caption{The $\mathrm{MR}^{-2}$ of RepGT and RepBox Losses and their improvements with different smooth parameters $\sigma$ on the validation set of CityPersons.}
\label{tab:citypersons_reasonable_ablation_sigma}
\end{table}

\begin{table}[!tbp]
\begin{center}
\setlength{\tabcolsep}{5pt}
\begin{tabular}{|c|c|c|c|c|c|}
\hline
 $\alpha$ (RepGT) & 0.3 & 0.4 & 0.5 & 0.6 & 0.7 \\
\hline
$\beta$ (RepBox) & 0.7 & 0.6 & 0.5 & 0.4 & 0.3 \\
\hline
$\mathrm{MR}^{-2}$ & 13.9 & 13.9 & {\bf 13.2} & 13.3 & 14.1  \\
\hline
\end{tabular}
\end{center}
\caption{We balance the RepGT and RepBox Losses by adjusting the weights $\alpha$ and $\beta$. Empirically, $\alpha=0.5$ and $\beta=0.5$ yields the best performance. The results are obtained on CityPersons validation subset.}
\vspace{-0.2cm}
\label{tab:citypersons_reasonable_ablation_combination}
\end{table}

\begin{figure}[t]
\includegraphics[width=0.42\textwidth]{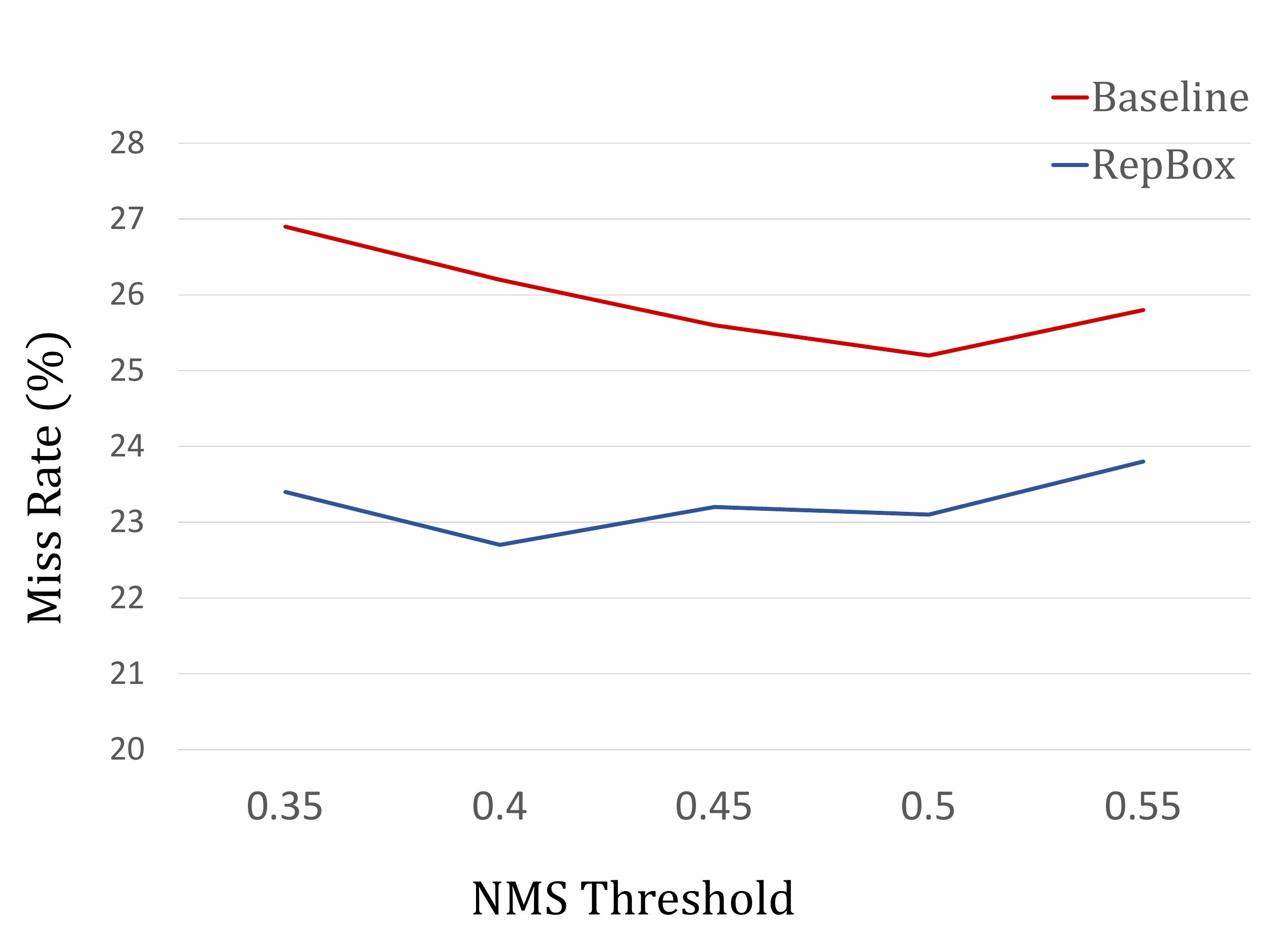}
\caption{Results with RepBox Loss across various NMS thresholds at $\mathrm{FPPI}=10^{-2}$. The curve of RepBox is smoother than that of baseline, indicating it is less sensitive to the NMS threshold.}
\vspace{-0.2cm}
\label{fig:RepBox_nms}
\end{figure}

\begin{table*}[!ht]
\begin{center}
 		\begin{tabular}{c|cccc|c|ccc}
 		\hline\hline
 		Method & +RepGT & +RepBox & +Segmentation & Scale & Reasonable & Heavy & Partial & Bare\\
 		\hline\hline
 		\multirow{3}{*}{Zhang \etal~\cite{zhang2017citypersons}} & & & & $\times{}1$ & 15.4 & 55.0 & 18.9 & 9.3 \\
 		& & & $\checkmark$ & $\times{}1$ & 14.8 & - & - & - \\
 		& & & & $\times{}1.3$ & 12.8 & - & - & -\\
 		\hline
 		Baseline & & & & $\times{}1$ & 14.6 & 60.6 & 18.6 & 7.9 \\
 		\hline
 		RepLoss & $\checkmark$ & & & $\times{}1$ & 13.7 & 57.5 & 17.3 & \textcolor{green}{7.2} \\
 		& & $\checkmark$ & & $\times{}1$ & 13.7 & 59.1 & 17.2 & 7.8 \\
 		& $\checkmark$ & $\checkmark$  & & $\times{}1$ & \textcolor{green}{13.2} & \textcolor{green}{56.9} & \textcolor{green}{16.8} & 7.6 \\
 		& $\checkmark$ & $\checkmark$  & & $\times{}1.3$ & \textcolor{blue}{11.6} & \textcolor{blue}{55.3}  & \textcolor{blue}{14.8} & \textcolor{blue}{7.0} \\
 		& $\checkmark$ & $\checkmark$  & & $\times{}1.5$ & \textcolor{red}{10.9} & \textcolor{red}{52.9}  & \textcolor{red}{13.4} & \textcolor{red}{6.3} \\
 		\end{tabular}
 	\end{center}
 	\caption{Pedestrian detection results using RepLoss evaluated on the CityPersons~\cite{zhang2017citypersons}. Models are trained on train set and tested on validation set. We use ResNet-50 as our back-bone architecture. The best 3 results are highlighted in red, blue and green, respectively.}
 	\label{tab:citypersons_results}
\end{table*}

\subsection{Ablation Study}
\label{subsec:ablationexperiments}

\myparagraph{RepGT Loss.} In Table~\ref{tab:citypersons_reasonable_ablation_sigma}, we report the results of RepGT Loss with different parameter $\sigma$ for $\mathrm{Smooth}_{ln}$ loss. When set $\sigma$ as $1.0$, adding RepGT Loss yields the best performance of $13.7$ $\mathrm{MR}^{-2}$ in terms of reasonable evaluation setup. It outperforms the baseline with an improvement of $0.9$ $\mathrm{MR}^{-2}$. Setting $\sigma=1$ that means we directly sum over $-\ln{(1-IoG)}$ with no smooth at all, similar to the loss function used in IoU Loss~\cite{yu2016unitbox}.

We also provide comparisons on missed detections and false positives between RepGT and baseline. In Figure~\ref{fig:crowd_miss}, adding RepGT Loss effectively decreases the number of missed detections in the reasonable-crowd subset. The curve of RepGT is consistently lower than that of baseline when the threshold of detection score is rather high, but two curves agree when the score is at $0.5$. The saturation points of curves are both at $\sim{}0.9$, also a commonly used threshold for real applications, where we reduce the quantity of missed detections by relatively $10\%$. In Figure~\ref{fig:crowd_fp}, false positives produced by RepGT Loss due to crowd occlusion cover less proportion than the baseline detector. This demonstrates that RepGT Loss is effective on reducing missed detections and false positives in crowd scenes.

\myparagraph{RepBox Loss.} For RepBox Loss, we experiment with a different smooth parameter $\sigma$, reported in the fourth line of Table~\ref{tab:citypersons_reasonable_ablation_sigma}. When setting $\sigma$ as $0$, RepBox Loss yields the best performance of $13.7$ $\mathrm{MR}^{-2}$, on par with RepGT with $\sigma=1.0$. Setting $\sigma$ as 0 means we completely smooth a $\ln$ function into a linear function and sum over IoU. We conjure that RepBox Loss tends to have more outliers than RepGT Loss since predicted boxes are much denser than ground-truth boxes.

\begin{figure}[!tbp]
\centering
\includegraphics[width=0.46\textwidth]{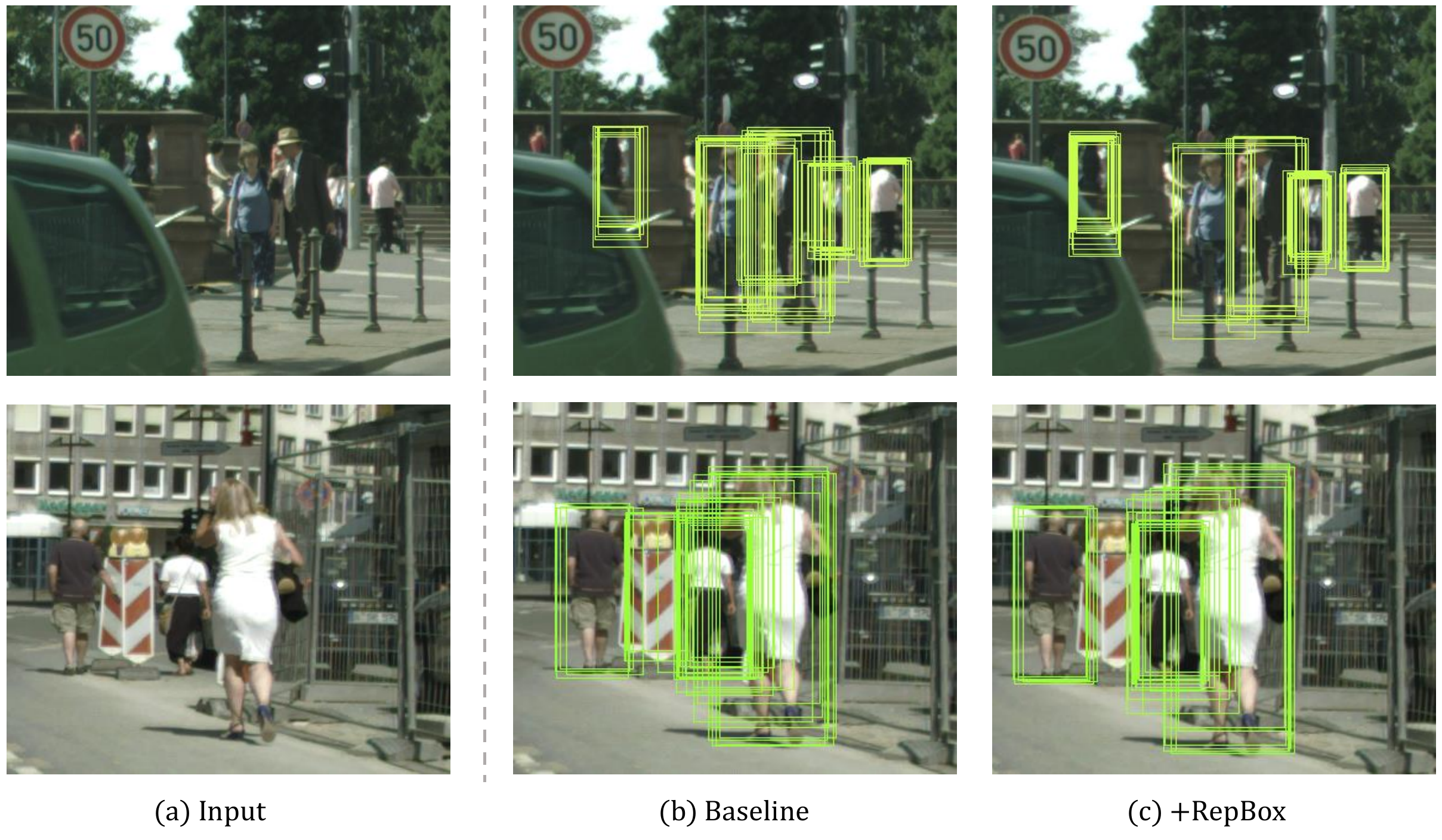}
\caption{Visualized comparison of predicted bounding boxes before NMS of baseline and RepBox. In the results of RepBox, there are fewer predictions lying in between two adjacent ground-truths, which is desirable in crowd scenes. More examples are shown in supplementary material.}
\vspace{-0.4cm}
\label{fig:vis_box}
\end{figure}

As mentioned in Section~\ref{sec:intro}, detectors in crowd scenes are sensitive to the NMS threshold. A high NMS threshold may lead to more false positives, while a low NMS threshold may lead to more missed detections. In Figure~\ref{fig:RepBox_nms} we show our results with RepBox Loss across various NMS thresholds at $\mathrm{FPPI}=10^{-2}$. In general, the performance of detector with RepBox Loss is smoother than baseline. It is worth noting that at the NMS threshold of $0.35$, the gap between baseline and RepBox is $3.5$ points, indicating that the latter is less sensitive to NMS threshold. Through visualization in Figure~\ref{fig:vis_box}, there are fewer predictions lying in between two adjacent ground-truths of RepBox, which is desirable in crowd scenes. More examples are shown in supplementary material.

\myparagraph{Balance of RepGT and RepBox}
The introduced RepGT and RepBox Loss help detectors do better in crowd scenes when added alone, but we have yet studied how to balance these two losses.  Table~\ref{tab:citypersons_reasonable_ablation_combination} shows our results with different settings of $\alpha$ and $\beta$. Empirically, $\alpha=0.5$ and $\beta=0.5$ yields the best performance.

\begin{table}[!tbp]
\begin{center}
\setlength{\tabcolsep}{5.5pt}
\begin{tabular}{c|c|c}
\toprule[1pt]
\multirow{2}{*}{Method} & \multicolumn{2}{c}{Reasonable}  \\
\cline{2-3}
& IoU=0.5 & IoU=0.75  \\
\midrule
Zhang \etal~\cite{zhang2017citypersons} & 5.8 & 30.6  \\
Mao \etal~\cite{mao2017can} & 5.5 & 43.4  \\
Zhang \etal~\cite{zhang2017citypersons}* & 5.1 & 25.8  \\
\midrule
Baseline & 5.6 & 28.7 \\
+RepGT & 5.0 & 27.1 \\
+RepBox & 5.3 & \textcolor{blue}{26.2} \\
+RepGT \& RepBox & \textcolor{blue}{5.0} & 26.3  \\
+RepGT \& RepBox* & \textcolor{red}{4.0} & \textcolor{red}{23.0}   \\
\bottomrule[1pt]
\end{tabular}
\end{center}
\caption{Results on Calech-USA test set (reasonable), evaluated on the new annotations~\cite{zhang2016far}. On a strong baseline, we further improve the state-of-the-art to a remarkable $4.0$ $\mathrm{MR}^{-2}$ under $0.5$ IoU threshold. The consistent gain when increasing IoU threshold to $0.75$ demonstrates effectiveness of repulsion loss. *: indicates pre-training network using CityPersons dataset.}
\vspace{-0.3cm}
\label{tab:caltech_reasonable}
\end{table}

\subsection{Comparisons with State-of-the-art Methods}
\label{subsec:compareison_with_stateoftheart}
To demonstrate our effectiveness under different occlusion levels, we divide the reasonable subset (occlusion $\leq$ 35\%) into the {\it reasonable-partial} subset (10\% $<$ occlusion $\leq$ 35\%), denoted as Partial subset, and the {\it reasonable-bare} subset (occlusion $\leq$ 10\%), denoted as Bare subset. For annotations whose occlusion is above $35\%$ (not in the reasonable set), we denote them as Heavy subset. Table~\ref{tab:citypersons_results} summarizes our results on CityPersons. In general, RepGT Loss and RepBox Loss show improvement across all evaluation subsets. Combined together, our proposed repulsion loss achieves $13.2$ $\mathrm{MR}^{-2}$, which is an absolute $1.4$-point improvement over our baseline. In terms of different occlusion levels, performance with RepLoss on the Heavy subset is boosted by a remarkably large margin of $3.7$ points, and on the Partial subset by a relatively smaller margin of $1.8$ points, while causing non-obvious improvement on the Bare subset. It is in accordance with our intention that RepLoss is specifically designed to address the occlusion problem.

We also evaluate RepLoss on new Caltech-USA dataset. Results are shown in Table~\ref{tab:caltech_reasonable}. On a strong reference, RepLoss achieves $\mathrm{MR^{-2}}$ of $5.0$ at $.5$ IoU matching threshold and $26.3$ at $.75$ IoU matching threshold. The consistent and even larger gain when increasing IoU threshold demonstrates the ability of our framework to handle occlusion problem, for it that occlusion is known for its tendency of being more sensitive at a higher matching threshold. Result curves are shown in Figure~\ref{fig:caltech_curve}. 

\iffalse
\begin{figure}[t]
\centering
\includegraphics[width=0.46\textwidth]{figures/caltech_curve.pdf}
\caption{Comparisons with state-of-the-art methods on the Caltech test set (reasonable subset) using new annotations.}
\vspace{-0.5cm}
\label{fig:caltech_curve}
\end{figure}
\fi

\begin{figure}[!t]
\centering
\includegraphics[width=0.43\textwidth]{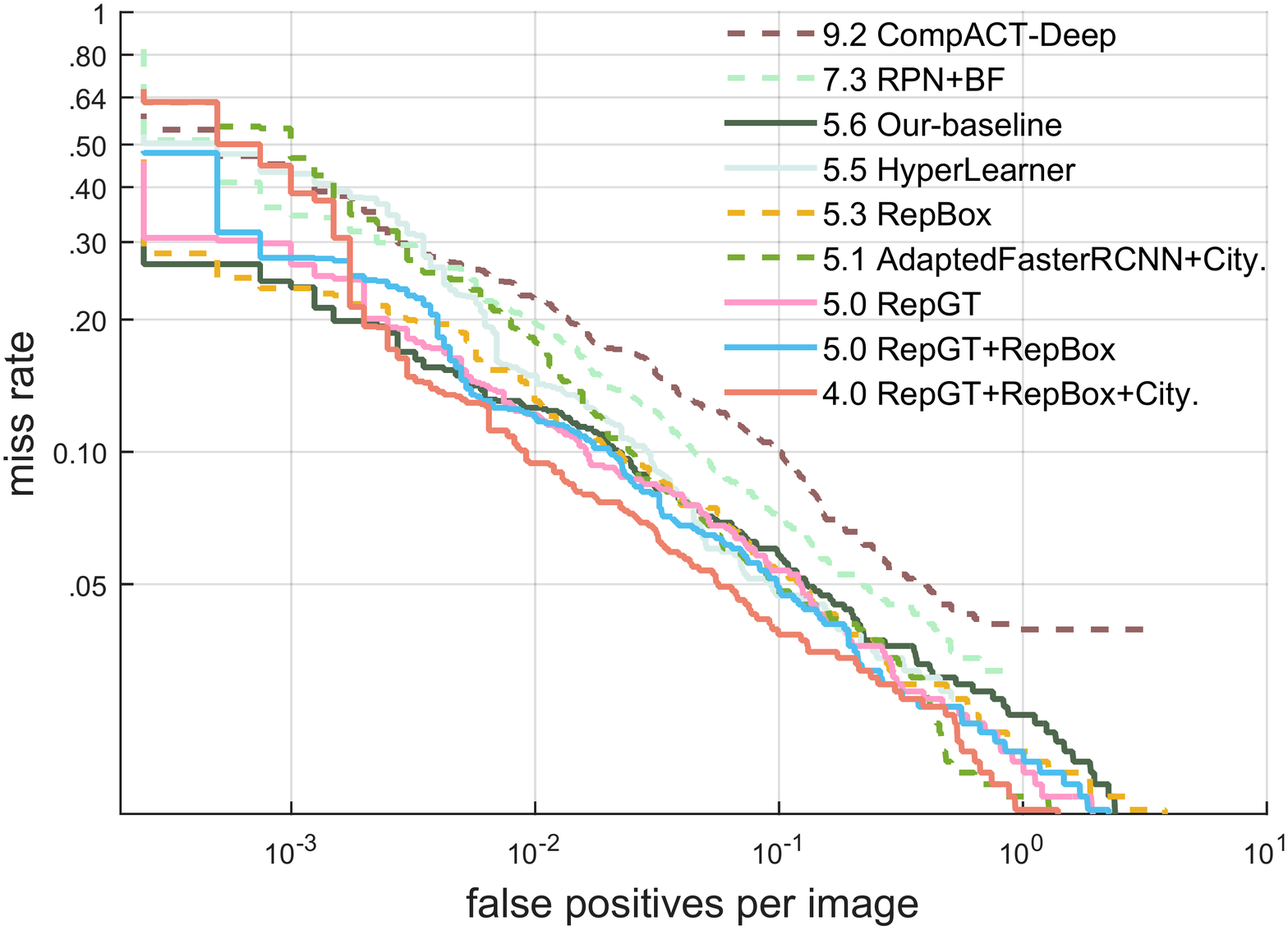}
\caption{Comparisons with state-of-the-art methods on the new Caltech test subset.}
\vspace{-0.6cm}
\label{fig:caltech_curve}
\end{figure}

\section{Extensions: General Object Detection}
Our RepLoss is a generic loss function for object detection in crowd scenes and can be used in applications other than pedestrian detection. In this section, we apply the repulsion loss to general object detection.

We conduct our experiments on the PASCAL VOC dataset~\cite{everingham2010pascal}, a common evaluation benchmark for general object detection. This dataset consists of over 20 object categories. Standard evaluation metric for VOC dataset is mean Average Precision (mAP) over all categories. We adopt the vanilla Faster R-CNN~\cite{NIPS2015_5638} framework, using ImageNet-pretrained ResNet-101~\cite{he2016deep} as the backbone. The NMS threshold is set as $0.3$. The model is trained on the train and validation subsets of PASCAL VOC 2007 and PASCAL VOC 2012, and is evaluated on the test subset of PASCAL VOC 2007. Our re-implemented baseline is better than original one by $3.4$ mAP.

Results are shown in Table~\ref{tab:voc}. The gain over the entire dataset is not significant. Nevertheless, when evaluated on the crowd subset (objects have intra-class IoU greater than 0.1), RepLoss outperforms the baseline by $2.1$ mAP. These results demonstrate that our method is generic and can be extended to general object detection.

\begin{table}[!tbp]
\begin{center}
\setlength{\tabcolsep}{5pt}
\begin{tabular}{c|c|c}
\toprule[1pt]
Method & mAP & mAP on Crowd \\
\hline
Faster R-CNN~\cite{he2016deep} & 76.4 & - \\
\hline 
Faster R-CNN (\textit{ReIm}) & 79.5 & 38.7 \\
+ RepGT & {\bf 79.8} & {\bf 40.8} \\
\bottomrule[1pt]
\end{tabular}
\end{center}
\caption{General object detection results evaluated on PASCAL VOC 2007~\cite{everingham2010pascal} benchmark. {\it ReIm} is our re-implemented Faster R-CNN. Crowd subset contains ground-truth objects who has overlaps above $0.1$ IoU region with at least another ground-truth object of the same category. Our RepGT Loss outperforms baseline by $2.1$ mAP on crowd subset.}
\vspace{-0.3cm}
\label{tab:voc}
\end{table}

\section{Conclusion}
In this paper, we have carefully designed the repulsion loss (RepLoss) for pedestrian detection, which improves detection performance, particularly in crowd scenes. The main motivation of the repulsion loss is that the attraction-by-target loss alone may not be sufficient for training an optimal detector, and repulsion-by-surrounding can be very beneficial. 

To implement the repulsion energy, we have introduced two types of repulsion losses. We have achieved the best reported performance on two popular datasets: Caltech and CityPersons. Significantly, our result on CityPersons without using pixel annotation outperforms the previously best result~\cite{zhang2017citypersons} that uses pixel annotation by about 2\%. Detailed experimental comparison have demonstrated the value of the proposed RepLoss, which improves detection accuracy by a large margin in occlusion scenarios. Results on generic object detection (PASCAL VOC) further show its usefulness. We expect wide application of the proposed loss in many other object detection tasks.

{\small
\bibliographystyle{ieee}
\bibliography{reploss}
}

{\small
\bibliographystyle{ieee}
\bibliography{reploss}

\begin{thebibliography}{10}\itemsep=-1pt

\bibitem{cai2016unified}
Z.~Cai, Q.~Fan, R.~S. Feris, and N.~Vasconcelos.
\newblock A unified multi-scale deep convolutional neural network for fast
  object detection.
\newblock In {\em European Conference on Computer Vision}, pages 354--370.
  Springer, 2016.

\bibitem{cordts2016cityscapes}
M.~Cordts, M.~Omran, S.~Ramos, T.~Rehfeld, M.~Enzweiler, R.~Benenson,
  U.~Franke, S.~Roth, and B.~Schiele.
\newblock The cityscapes dataset for semantic urban scene understanding.
\newblock In {\em Proceedings of the IEEE Conference on Computer Vision and
  Pattern Recognition}, pages 3213--3223, 2016.

\bibitem{dai2016r}
J.~Dai, Y.~Li, K.~He, and J.~Sun.
\newblock R-fcn: Object detection via region-based fully convolutional
  networks.
\newblock In {\em Advances in neural information processing systems}, pages
  379--387, 2016.

\bibitem{desai2011discriminative}
C.~Desai, D.~Ramanan, and C.~C. Fowlkes.
\newblock Discriminative models for multi-class object layout.
\newblock {\em International journal of computer vision}, 95(1):1--12, 2011.

\bibitem{dollar2014fast}
P.~Doll{\'a}r, R.~Appel, S.~Belongie, and P.~Perona.
\newblock Fast feature pyramids for object detection.
\newblock {\em IEEE Transactions on Pattern Analysis and Machine Intelligence},
  36(8):1532--1545, 2014.

\bibitem{dollar2009integral}
P.~Doll{\'a}r, Z.~Tu, P.~Perona, and S.~Belongie.
\newblock Integral channel features.
\newblock 2009.

\bibitem{dollar2009pedestrian}
P.~Doll{\'a}r, C.~Wojek, B.~Schiele, and P.~Perona.
\newblock Pedestrian detection: A benchmark.
\newblock In {\em Computer Vision and Pattern Recognition, 2009. CVPR 2009.
  IEEE Conference on}, pages 304--311. IEEE, 2009.

\bibitem{everingham2010pascal}
M.~Everingham, L.~Van~Gool, C.~K. Williams, J.~Winn, and A.~Zisserman.
\newblock The pascal visual object classes (voc) challenge.
\newblock {\em International journal of computer vision}, 88(2):303--338, 2010.

\bibitem{Girshick_2015_ICCV}
R.~Girshick.
\newblock Fast r-cnn.
\newblock In {\em The IEEE International Conference on Computer Vision (ICCV)},
  December 2015.

\bibitem{Girshick_2014_CVPR}
R.~Girshick, J.~Donahue, T.~Darrell, and J.~Malik.
\newblock Rich feature hierarchies for accurate object detection and semantic
  segmentation.
\newblock In {\em The IEEE Conference on Computer Vision and Pattern
  Recognition (CVPR)}, June 2014.

\bibitem{he2017mask}
K.~He, G.~Gkioxari, P.~Doll{\'a}r, and R.~Girshick.
\newblock Mask r-cnn.
\newblock In {\em The IEEE International Conference on Computer Vision (ICCV)},
  2017.

\bibitem{he2016deep}
K.~He, X.~Zhang, S.~Ren, and J.~Sun.
\newblock Deep residual learning for image recognition.
\newblock In {\em Proceedings of the IEEE conference on computer vision and
  pattern recognition}, pages 770--778, 2016.

\bibitem{Hosang_2017_CVPR}
J.~Hosang, R.~Benenson, and B.~Schiele.
\newblock Learning non-maximum suppression.
\newblock In {\em The IEEE Conference on Computer Vision and Pattern
  Recognition (CVPR)}, July 2017.

\bibitem{hosang2015taking}
J.~Hosang, M.~Omran, R.~Benenson, and B.~Schiele.
\newblock Taking a deeper look at pedestrians.
\newblock In {\em Proceedings of the IEEE Conference on Computer Vision and
  Pattern Recognition}, pages 4073--4082, 2015.

\bibitem{huang2015densebox}
L.~Huang, Y.~Yang, Y.~Deng, and Y.~Yu.
\newblock Densebox: Unifying landmark localization with end to end object
  detection.
\newblock {\em arXiv preprint arXiv:1509.04874}, 2015.

\bibitem{krizhevsky2012imagenet}
A.~Krizhevsky, I.~Sutskever, and G.~E. Hinton.
\newblock Imagenet classification with deep convolutional neural networks.
\newblock In {\em Advances in neural information processing systems}, pages
  1097--1105, 2012.

\bibitem{li2017scale}
J.~Li, X.~Liang, S.~Shen, T.~Xu, J.~Feng, and S.~Yan.
\newblock Scale-aware fast r-cnn for pedestrian detection.
\newblock {\em IEEE Transactions on Multimedia}, 2017.

\bibitem{li2017fully}
Y.~Li, H.~Qi, J.~Dai, X.~Ji, and Y.~Wei.
\newblock Fully convolutional instance-aware semantic segmentation.
\newblock In {\em IEEE Conf. on Computer Vision and Pattern Recognition
  (CVPR)}, pages 2359--2367, 2017.

\bibitem{Lin_2017_CVPR}
T.-Y. Lin, P.~Doll{\'a}r, R.~Girshick, K.~He, B.~Hariharan, and S.~Belongie.
\newblock Feature pyramid networks for object detection.
\newblock In {\em The IEEE Conference on Computer Vision and Pattern
  Recognition (CVPR)}, 2017.

\bibitem{Lin_2017_ICCV}
T.-Y. Lin, P.~Goyal, R.~Girshick, K.~He, and P.~Doll{\'a}r.
\newblock Focal loss for dense object detection.
\newblock In {\em The IEEE International Conference on Computer Vision (ICCV)},
  2017.

\bibitem{mao2017can}
J.~Mao, T.~Xiao, Y.~Jiang, and Z.~Cao.
\newblock What can help pedestrian detection?
\newblock In {\em The IEEE Conference on Computer Vision and Pattern
  Recognition (CVPR)}, 2017.

\bibitem{nam2014local}
W.~Nam, P.~Doll{\'a}r, and J.~H. Han.
\newblock Local decorrelation for improved detection.
\newblock {\em arXiv preprint arXiv:1406.1134}, 2014.

\bibitem{ouyang2012discriminative}
W.~Ouyang and X.~Wang.
\newblock A discriminative deep model for pedestrian detection with occlusion
  handling.
\newblock In {\em Computer Vision and Pattern Recognition (CVPR), 2012 IEEE
  Conference on}, pages 3258--3265. IEEE, 2012.

\bibitem{NIPS2015_5638}
S.~Ren, K.~He, R.~Girshick, and J.~Sun.
\newblock Faster r-cnn: Towards real-time object detection with region proposal
  networks.
\newblock In C.~Cortes, N.~D. Lawrence, D.~D. Lee, M.~Sugiyama, and R.~Garnett,
  editors, {\em Advances in Neural Information Processing Systems 28}, pages
  91--99. Curran Associates, Inc., 2015.

\bibitem{shrivastava2016training}
A.~Shrivastava, A.~Gupta, and R.~Girshick.
\newblock Training region-based object detectors with online hard example
  mining.
\newblock In {\em Proceedings of the IEEE Conference on Computer Vision and
  Pattern Recognition}, pages 761--769, 2016.

\bibitem{simonyan2014very}
K.~Simonyan and A.~Zisserman.
\newblock Very deep convolutional networks for large-scale image recognition.
\newblock {\em arXiv preprint arXiv:1409.1556}, 2014.

\bibitem{tian2015deep}
Y.~Tian, P.~Luo, X.~Wang, and X.~Tang.
\newblock Deep learning strong parts for pedestrian detection.
\newblock In {\em Proceedings of the IEEE international conference on computer
  vision}, pages 1904--1912, 2015.

\bibitem{yang2015convolutional}
B.~Yang, J.~Yan, Z.~Lei, and S.~Z. Li.
\newblock Convolutional channel features.
\newblock In {\em Proceedings of the IEEE international conference on computer
  vision}, pages 82--90, 2015.

\bibitem{yu2016unitbox}
J.~Yu, Y.~Jiang, Z.~Wang, Z.~Cao, and T.~Huang.
\newblock Unitbox: An advanced object detection network.
\newblock In {\em Proceedings of the 2016 ACM on Multimedia Conference}, pages
  516--520. ACM, 2016.

\bibitem{zhang2016faster}
L.~Zhang, L.~Lin, X.~Liang, and K.~He.
\newblock Is faster r-cnn doing well for pedestrian detection?
\newblock In {\em European Conference on Computer Vision}, pages 443--457.
  Springer, 2016.

\bibitem{zhang2016far}
S.~Zhang, R.~Benenson, M.~Omran, J.~Hosang, and B.~Schiele.
\newblock How far are we from solving pedestrian detection?
\newblock In {\em IEEE Conference on Computer Vision and Pattern Recognition}.
  IEEE Computer Society, 2016.

\bibitem{zhang2015filtered}
S.~Zhang, R.~Benenson, and B.~Schiele.
\newblock Filtered channel features for pedestrian detection.
\newblock In {\em 2015 IEEE Conference on Computer Vision and Pattern
  Recognition (CVPR)}, pages 1751--1760. IEEE, 2015.

\bibitem{zhang2017citypersons}
S.~Zhang, R.~Benenson, and B.~Schiele.
\newblock Citypersons: A diverse dataset for pedestrian detection.
\newblock In {\em The IEEE Conference on Computer Vision and Pattern
  Recognition (CVPR)}, 2017.

\bibitem{zhou2017multi}
C.~Zhou and J.~Yuan.
\newblock Multi-label learning of part detectors for heavily occluded
  pedestrian detection.
\newblock In {\em Proceedings of the IEEE Conference on Computer Vision and
  Pattern Recognition}, pages 3486--3495, 2017.

\end{thebibliography}
}

\clearpage

\part*{Supplementary material}

\begin{figure*}[!tbp]
\centering
\includegraphics[width=0.8\textwidth]{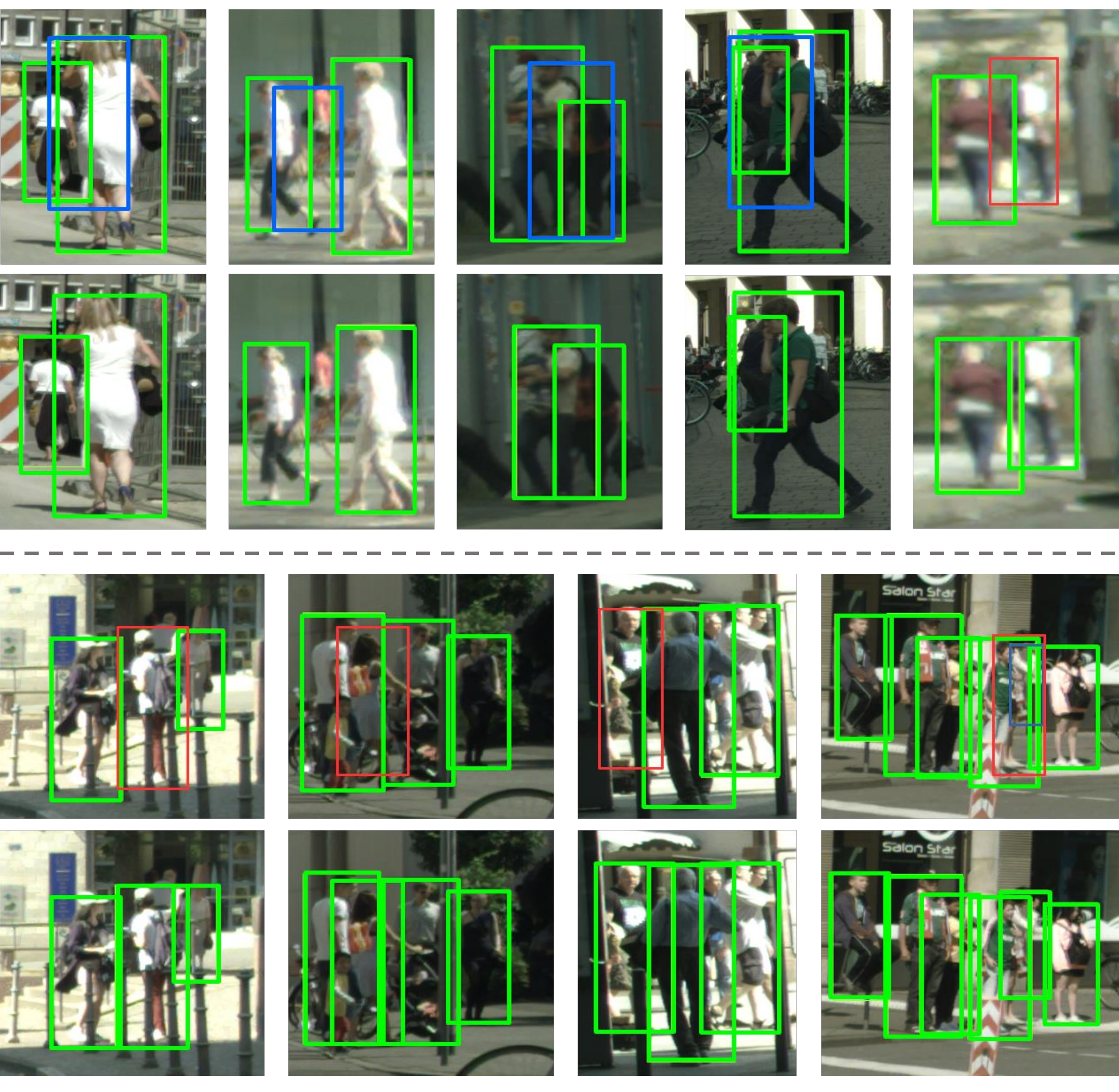}
\caption{Comparison of baseline and RepLoss. The blue bounding boxes represent false positives, and the red ones represent the missed detections. On two sides of the grey dashed line, samples on the first row of each side are predictions of our baseline, while samples on the second row of each side are the predictions after adding the RepLoss.}
\label{fig:supp_compare}
\end{figure*}

\textbf{A. The examples of missed detections and false positives} 

In Figure~\ref{fig:supp_compare} we show more examples of missed detections and false positives before and after applying RepLoss. The blue bounding boxes represent false positives, and the red ones represent the missed detections. The examples above the grey dashed line are to demonstrate the effectiveness of proposed RepLoss on {\it eliminating false positives}, while those bolow the grey dashed line are to demonstrate the effectiveness of proposed RepLoss on {\it detecting more missed pedestrians.}
\par

\begin{figure*}
\centering
\includegraphics[width=0.98\textwidth]{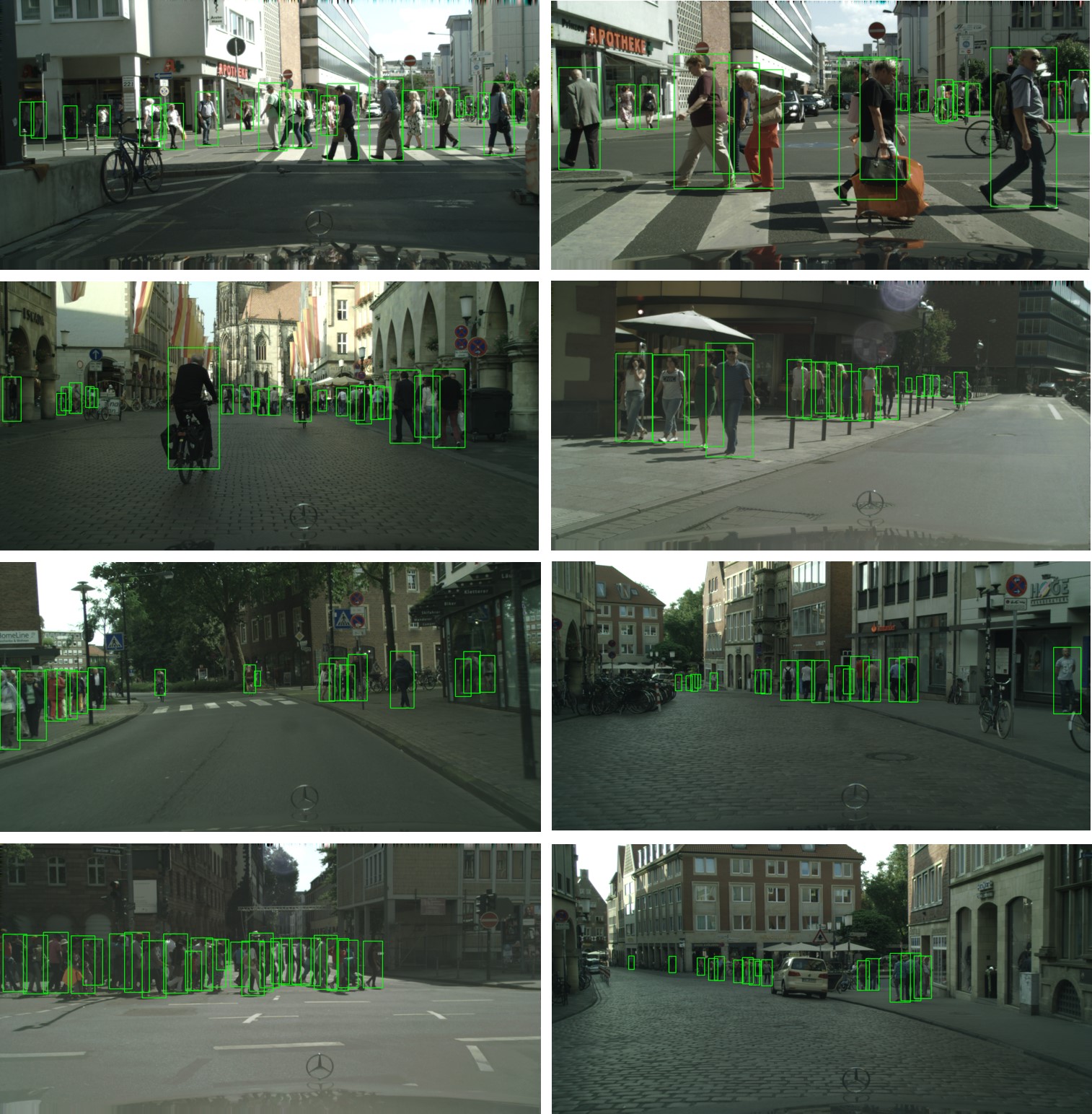}
\caption{More examples on CityPersons dataset. Green bounding boxes are predicted pedestrians whose score ([0, 1.0]) is greater than 0.8.}
\label{fig:supp_show}
\end{figure*}

\textbf{B. More examples on CityPersons}
In Figure~\ref{fig:supp_show}, we demonstrate more examples on challenging CityPersons dataset. Green bounding boxes are predicted pedestrians whose score (in range $\left[0, 1.0\right]$) is at a relatively high threshold (greater than 0.8 in this case).

\end{document}